\documentclass[letterpaper,10pt,conference]{IEEEtran}  

\usepackage{xparse} 
\usepackage{amsmath} 
\usepackage{amssymb}
\usepackage{amsfonts} 
\usepackage{dsfont}
\usepackage{bbm}
\usepackage{color}
\usepackage{verbatim}
\usepackage{multirow}
\usepackage[flushleft]{threeparttable}
\usepackage{graphicx}
\usepackage[font=small]{caption}
\usepackage{balance}

\usepackage[bookmarks=false,hidelinks]{hyperref}
\usepackage{cleveref}
\usepackage[font=small,position=b]{subcaption}
\usepackage{color}
\definecolor{mygray}{gray}{0.4}
\usepackage{float}
\usepackage{stfloats}
\usepackage{caption}
\usepackage{cuted}
\usepackage{mathrsfs}

\voffset=\baselineskip

\IEEEoverridecommandlockouts
\IEEEpubid{\makebox[\columnwidth]
	{\hspace{0.4em}978-1-5386-5635-8/18/\$31.00~\copyright{\hspace{0.15em}2018} IEEE.\hfill}%
	\hspace{\columnsep}\makebox[\columnwidth]{ }}

\usepackage{fancyhdr}
\fancypagestyle{firstpage}{
	\fancyhf{}
	\fancyhead[C]{\textcolor{mygray}{2018 International Conference on Indoor Positioning and Indoor Navigation (IPIN), 24-27 September 2018, Nantes, France}}
	\makeatletter
	\fancyfoot[L]{%
		\normalfont\footnotesize%
		\raisebox{\footskip}[1.5ex]{%
			\makebox[\columnwidth]{~~~}}}
}

\usepackage{url}
\usepackage{framed}

\usepackage{booktabs}
\usepackage{arydshln}

\newcommand\T{\rule{0pt}{2.6ex}}        
\newcommand\B{\rule[-1.2ex]{0pt}{0pt}}  

\newcommand{\mbf}[1]{\mathbf{#1}}

\DeclareMathAlphabet{\mbfh}{OML}{cmm}{b}{it}

\newcommand{\norm}[1]{\left\Vert#1\right\Vert}

\newcommand{\bbm}{\begin{bmatrix}}
\newcommand{\ebm}{\end{bmatrix}}

\newcommand{\SO}[1]{SO(#1)}

\title{\LARGE \bf LSTM-Based Zero-Velocity Detection for Robust Inertial Navigation}
\author{Brandon Wagstaff and Jonathan Kelly\thanks{All authors are with the Space \& Terrestrial Autonomous Robotic Systems (STARS) Laboratory at the University of Toronto Institute for Aerospace Studies (UTIAS), Toronto, Canada. Email: \texttt{brandon.wagstaff@robotics.utias.utoronto.ca}, \texttt{jonathan.kelly@robotics.utias.utoronto.ca}.}}

\begin{document}
\maketitle 
\thispagestyle{firstpage}

\begin{abstract}
We present a method to improve the accuracy of a zero-velocity-aided inertial navigation system (INS) by replacing the standard zero-velocity detector with a long short-term memory (LSTM) neural network.  While existing threshold-based zero-velocity detectors are not robust to varying motion types, our learned model accurately detects stationary periods of the inertial measurement unit (IMU) despite changes in the motion of the user. Upon detection, zero-velocity pseudo-measurements are fused with a dead reckoning motion model in an extended Kalman filter (EKF). We demonstrate that our LSTM-based zero-velocity detector, used within a zero-velocity-aided INS, improves zero-velocity detection during human localization tasks. Consequently, localization accuracy is also improved.

Our system is evaluated on more than 7.5 km of indoor pedestrian locomotion data, acquired from five different subjects. We show that 3D positioning error is reduced by over 34\% compared to existing fixed-threshold zero-velocity detectors for walking, running, and stair climbing motions. Additionally, we demonstrate how our learned zero-velocity detector operates effectively during crawling and ladder climbing. Our system is calibration-free (no careful threshold-tuning is required) and operates consistently with differing users, IMU placements, and shoe types, while being compatible with any generic zero-velocity-aided INS.
 \end{abstract}

\section{Introduction}

Reliable localization is vital in situations such as emergency response, during which a continuous position estimate for each first responder (e.g., a firefighter) is needed to ensure safety and coordinate activity. While Global Navigation Satellite System (GNSS) receivers are commonly used in outdoor environments, they fail to produce accurate localization information within buildings due to signal absorption and spurious reflections. Instead, body-mounted inertial measurement units (IMUs) can be used to record the movements of individuals, providing an estimate of their motion relative to a known origin. Unlike infrastructure-dependent localization systems, body-mounted IMUs are lightweight and can be rapidly and easily deployed.

Traditionally, an IMU is utilized as part of an inertial navigation system (INS), where the linear acceleration and angular velocity data from the unit are integrated to yield position updates over time. All IMUs are subject to drift, however; there are a number of possible ways to combat error accumulation caused by drift. An attractive, infrastructure-free technique is to mount the IMU on the foot of an individual---pseudo-measurements of the velocity state can then be obtained during `midstance', that is, the portion of the human gait when the foot is flat on the ground and stationary relative to the surface. By incorporating these pseudo-measurements into the INS, dead reckoning is required only during the intervals between footfalls, instead of along an entire trajectory (in the latter case, position error  grows cubically with time \cite{ZVDetect}). Therefore, an accurate zero-velocity-aided INS is contingent upon accurate zero-velocity detection. 

A known limitation of standard zero-velocity detectors is their threshold-based activation---given a fixed threshold, the detectors fail to perform reliably across a variety of gait motions. Recent approaches aim to improve zero-velocity detection during common motions by implementing adaptive techniques that are dependent on velocity \cite{Sensors2016,context-adaptive} or gait frequency \cite{Sensors2016_2}. However, modelling zero-velocity detection during motions such as stair climbing and crawling, while maintaining accurate midstance detection during walking and running, is fundamentally challenging \cite{Rantakokko:2012}. In past work \cite{Wagstaff:2017}, we showed that motion classification can be useful when combined with a series of motion-specific detectors, as this allows each detector to be optimized for a particular motion type. In this work, we relax the requirement for the human motion to be discretely classified (a priori) into a small set of motion types, as this is a simplification that inevitably reduces the detector's ability to perform well during transitions and for `undefined' motions.  The complexity of human movement requires a sophisticated zero-velocity detection model, which is difficult to hand-craft: we postulate that a more prudent solution is to take a data-driven approach and learn a gait representation that can be used for zero-velocity detection.
\begin{figure}[t]
	\small
	\centering	
	\includegraphics[width=0.9\columnwidth]{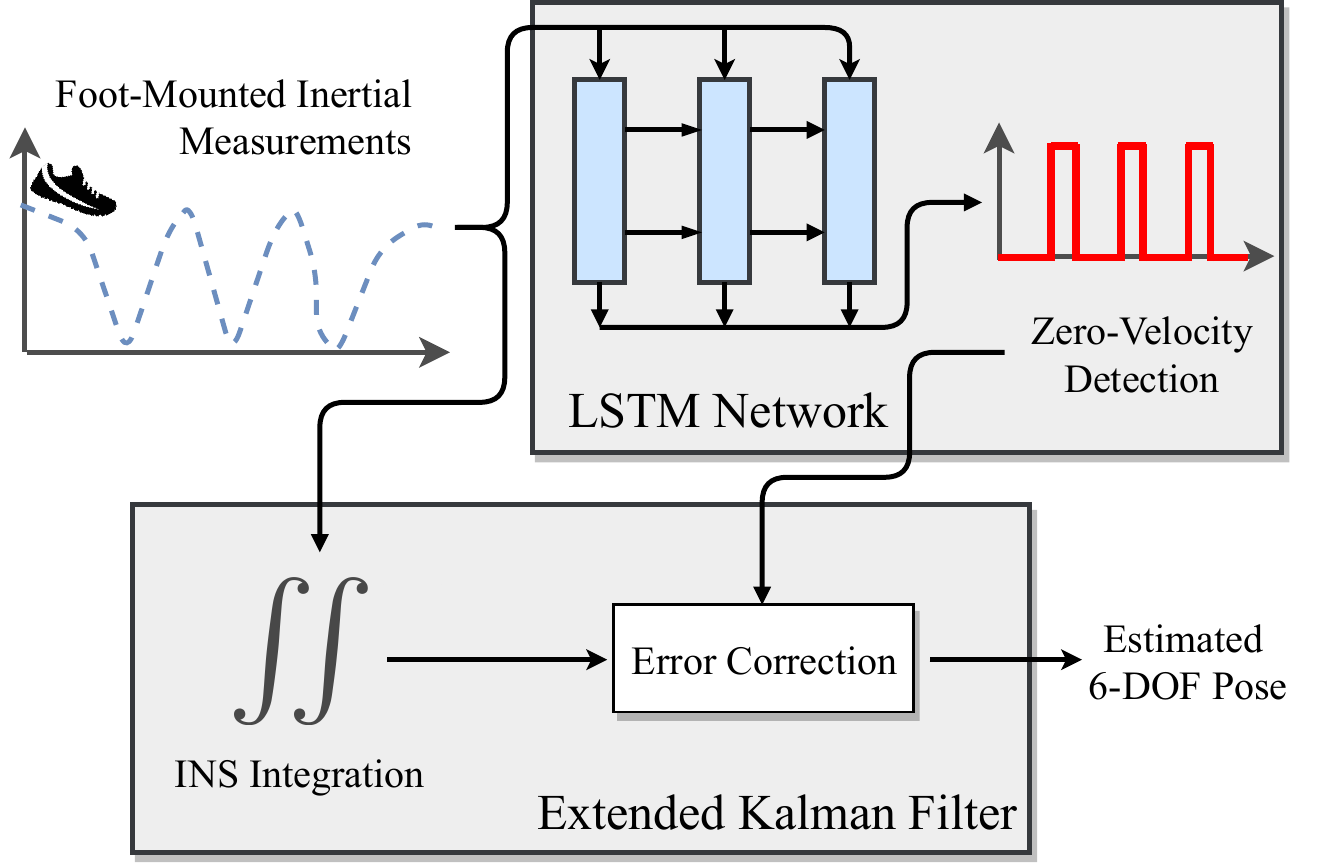}
	\vspace{1mm}
	\caption{Our proposed localization system, incorporating an LSTM-based zero-velocity detector.  Zero-velocity measurements are fused with a dead-reckoning motion model in an extended Kalman filter to significantly reduce error growth over time.}
	\vspace{-5mm}
	\label{fig:system}
\end{figure}

Our approach leverages the recent success of modern machine learning methods, by replacing the traditional zero-velocity detector with a long short-term memory (LSTM) neural network \cite{Hochreiter:1997} that outputs binary zero-velocity classifications from raw IMU data. The network is trained with accurate ground truth zero-velocity labels for a variety of human gait motions, and in turn is able to achieve high accuracy across a  range of motion types. Additionally, our system (illustrated in \Cref{fig:system}) operates consistently with different users, IMU placements, and shoe types, all without retraining.    

\vspace{2mm}
\section{Related Work}
\label{sec:related-work}

Inertial navigation systems rely on linear acceleration and angular velocity measurements to estimate the pose (position, velocity, orientation) of an IMU relative to a known origin. Integration of angular velocity measurements drives the orientation estimate, while integration of the gravity-subtracted linear acceleration measurements drives the velocity (single integration) and position (double integration) estimates. In a zero-velocity-aided INS, error accumulation originating from the integration of noisy IMU readings is reduced by introducing prior knowledge about the system, specifically that the IMU is frequently (approximately) stationary. Upon detection of a zero-velocity event, the pseudo-measurement of the velocity state can be fused with the dead reckoning motion model in an extended Kalman filter (EKF) \cite{Foxlin:2005} or another Bayesian filter.

Zero-velocity detection during midstance has been well-studied by Skog et al.\ \cite{zupteval,ZVDetect} and Olivares et al.\ \cite{Olivares:2012}, who both presented and characterized a range of threshold-based detection methods. Typically, zero-velocity detection is achieved through a likelihood ratio test (LRT) that indicates whether the IMU is stationary given the current IMU readings \cite{ZVDetect}. Any IMU reading that causes the LRT to fall below a specified threshold is considered to be due to a stationary sensor. While a fixed threshold may provide near-optimal zero-velocity detection for a uniform motion type, humans are capable of performing a wide range of motions. Fixed-threshold detection fails when a user changes their motion type or intensity, since the LRT threshold is no longer optimal. There are two possible failure cases: 1) the threshold is set to a higher-than-optimal value, causing the LRT output to fall below the threshold while the foot is still moving (i.e., a false-positive detection), or 2) the threshold is set to a lower-than-optimal value, causing the LRT output to remain higher than the threshold while the foot \emph{is} stationary (i.e., a false-negative detection). Both cases result in error accumulation \cite{Nilsson:2012}.

Several improvements to the baseline fixed-threshold detectors have been made to address the difficulties above, by using adaptive algorithms that alter the LRT threshold online according to a range of hand-crafted heuristics. For example, thresholds have been set to be proportional to the user's velocity \cite{Sensors2016,context-adaptive}, or gait frequency \cite{Sensors2016_2}. In our prior work, we implemented a motion classifier that actively updated the LRT threshold such that it was optimal for a limited set of motion types \cite{Wagstaff:2017}.

The work that is closest in spirit to that presented herein is by Park et al.\ \cite{Park2016}, who used foot-mounted inertial sensors in their learned zero-velocity detection system. Two separate support vector machine (SVM) classifiers were trained: one to classify a user's motion type, and another to identify stationary periods given the current motion type. While the authors of \cite{Park2016} reported high detection accuracies, the evaluation was for walking motions only. Also, the detector was not evaluated within a zero-velocity-aided INS. Furthermore, the system in \cite{Park2016} requires that a moving average filter be applied to the SVM output in order to remove false-positive detections; this may erroneously remove correctly-predicted zero-velocity events for running users. Our approach differs from Park et al.\ in that we train a single model for zero-velocity detection that operates independently of the current motion type. This allows our model to generalize to motions that cannot easily be discretized into predefined classes.

We draw further inspiration from other implementations that use deep learning to process inertial data.  Hannink et al. \cite{Hannink:2018} train a deep convolutional neural network (CNN) to regress human stride length from foot-mounted inertial data. IONet \cite{Chen:2018} is an end-to-end learned INS that provides a continuous trajectory estimate directly from raw inertial data using an LSTM network. While this implementation fully replaces the filtering architecture with a learned model, we believe that zero-velocity detection is an integral part of the architecture and that an end-to-end method would have difficulty reproducing the accuracy of a zero-velocity-aided system. Instead, we simply replace an error-prone component of the system (i.e., the zero-velocity detector) with a learned model, without modifying the remainder of the INS.

\vspace{2mm}
\section{LSTM-based zero-velocity detection}
\label{sec:lstm-zv}

We discuss our system in four sections. In \Cref{sec:datacollection1} and \Cref{sec:datacollection2}, we describe our procedure for collecting training and test data for our learned zero-velocity detector. In \Cref{sec:lstm-architecture}, we define our LSTM network architecture. In \Cref{sec:lstm-training}, we outline the training process and give an overview of the training results.

For training (and for a portion of our testing), we used a MicroStrain 3DM-GX3-25 IMU operating at 200 Hz.  All test subjects wore their preferred pair of running shoes during our experiments, with the IMU mounted at approximately the centre of the right foot, directly under the shoe's laces.

\subsection{Zero-Velocity Ground Truth Labelling}
\label{sec:datacollection1}

We initially collected a large dataset for training of our zero-velocity detector. The dataset consisted of raw IMU measurements, where, on a per-sample basis at each timestep, a binary label of `stationary' or `moving' was assigned. Acquiring accurate labels is non-trivial due to the uncertainty regarding whether the foot is absolutely stationary during a dynamic motion. Additionally, factors such as motion type, IMU placement on the foot, shoe type, and ground surface will impact zero-velocity detection. While there is no standard method of labelling midstance, prior studies have generated labels manually \cite{Olivares:2012}, with a pressure sensor \cite{zupteval}, or using a LRT \cite{Park2016}.

In this work, we make the assumption that, for a given motion trial, the most accurate zero-velocity labels that can be acquired are those producing the lowest position error (relative to ground truth) when used within a foot-mounted INS. We chose to evaluate the outputs of several zero-velocity detectors, each of which was optimized to produce the lowest average root-mean-square error (ARMSE) over a motion trial. We labelled the inertial data with the output from the detector that was optimally tuned to produce the minimum ARMSE.

For each motion trial, we generated midstance estimates from five zero-velocity detectors, by selecting the detector threshold that minimized position error over the trial. We note that, within each motion trial, individuals ensured that their movements were as uniform as possible (maintaining a fixed motion type and intensity), such that a fixed-threshold detector was approximately optimal over the sequence. Had the individuals altered their motion type or intensity during the trial, a fixed threshold would no longer be optimal, and the output from the detectors would not be accurate enough to label our dataset.  We repeated this process for all motion trials. The five detectors are:

\subsubsection{Stance Hypothesis Optimal Estimation (SHOE) Detector}

The SHOE detector \cite{ZVDetect} is based on a generalized likelihood ratio test (GLRT) that indicates how likely it is that the IMU is moving. If the likelihood falls below a threshold, $\gamma$, the hypothesis that the IMU is stationary is accepted (meaning the specific force measured is strictly due to gravity and that the angular rotation rate is zero). The GLRT output, $T_k(\mathbf{a}, \boldsymbol{\omega})$, is defined as
\begin{equation}
\label{eq:shoe}
\begin{split}
y_k &= T_k(\mathbf{a}, \boldsymbol{\omega}) \leq \gamma \\
&= \frac{1}{W} \sum_{n=k}^{k+W-1} \left( \frac{1}{\sigma_a^2}\norm{ \mathbf{a}_n - g\frac{\bar{\mathbf{a}}}{\norm{\bar{\mathbf{a}}}}}^2 + \frac{1}{\sigma_\omega^2}\norm{\boldsymbol{\omega}_n}^2 \right) \leq \gamma.
\end{split}
\end{equation}
\noindent Here $\mbf a$, $\boldsymbol{\omega} \in\mathds{R}^{W\times3}$ are inertial measurements in the current window (of size $W$), $\sigma_a^2, \sigma_\omega^2$ are the variances of the specific force and angular rate measurements, $\mbf {\bar{a}}$ is the per-channel mean of the specific force samples in W, and $g$ is the magnitude of the local gravitational acceleration.\footnote{We note that $\gamma$ is the primary tuning parameter that has the largest effect on detection.  For our experiments, we tune the threshold while leaving the other parameters  fixed at: $W = 5$, $\sigma_a=9.8\times10^{-4}$, and $\sigma_\omega =8.726\times10^{-5}$.}

\subsubsection{Angular Rate Energy Detector (ARED)}

The ARED \cite{ZVDetect} is a simplification of the SHOE detector that omits the linear acceleration component.  The IMU is assumed to be stationary when $T_k(\boldsymbol{\omega})$ falls below a threshold $\gamma_\omega$,
\begin{equation}
\label{eq:ARED}
\begin{split}
y_k &= T_k(\boldsymbol{\omega}) \leq \gamma_\omega \\
&= \frac{1}{W} \sum_{n=k}^{k+W-1} \norm{\boldsymbol{\omega}_n}^2  \leq \gamma_\omega.
\end{split}.
\end{equation}
\subsubsection{Acceleration-Moving Variance Detector (AMVD)}

The AMVD \cite{ZVDetect} computes a measure of the accelerometer variance over the sample window. If the variance measure falls below a threshold, $\gamma_v$, the IMU is assumed to be stationary,
\begin{equation}
\label{eq:AMVD}
\begin{split}
y_k &= T_k(\boldsymbol{a}) \leq \gamma_v \\
&= \frac{1}{W} \sum_{n=k}^{k+W-1} \norm{\mbf{a}_n - \mbf{\bar{a}}}^2  \leq \gamma_v.
\end{split}
\end{equation}
\subsubsection{Memory-Based Graph Theoretic Detector (MBGTD)}

The MBGTD \cite{Olivares:2012} computes the distance between two distributions, which consist of the samples in window W split into sub-windows.  The average Euclidean distance $C_{i,j}$ between samples in the two sub-windows split at points $i$ and $j$ is
\begin{equation}
C_{i,j} = \frac{\sum_{i=k}^{j-1} \sum_{l=j}^{k+W} \norm{\mbf{a}_i - \mbf{a}_l}}{(j-i)(W-j+1)}.
\end{equation}
The quantity $C_{i,j}$ is computed for every possible frame split, and zero velocity is assumed when the maximum $C_{i,j}$ value falls below a threshold, $\gamma_m$,
\begin{equation}
y_k = \left(\max_{k \leq i < j \leq k+W} {C}_{i,j}\right) \leq \gamma_m.
\end{equation}
\subsubsection{VICON Midstance Detection}

The final method of zero-velocity labelling makes use of a VICON optical tracking system available in our laboratory. In the same manner as \cite{Wagstaff:2017}, we attached VICON markers to the IMU (on the shoe) and recorded both the inertial data and VICON position data  using the Robot Operating System (ROS). This resulted in an accurate position measurement at every IMU timestep. By numerically differentiating the VICON position data $\{\mbf p_k\}$, we computed foot velocities $\mbf v_k$, and applied a threshold $\gamma_v \in [0.02,0.4] $ to generate ground truth zero velocity events $y_k \in \{0,1\}$.\footnote{The upper threshold of 0.4 m/s appears to be large for detection of zero-velocity. However, our VICON tracking frame used markers that protruded off of the IMU, causing a small angular velocity of the foot at midstance to result in a larger velocity as measured by VICON.} 
\begin{equation}
y_k = \norm{\frac{d\mbf{p}_k}{dt_k}} \leq \gamma_v.
\end{equation}

\subsection{Data Collection}
\label{sec:datacollection2}

We collected our motion dataset within our VICON capture area, which is approximately $5 \times 5$ m in size. As a result, the motions typically consisted of circular trajectories within this space. Data were collected from a single test subject who performed 60 separate motion trials consisting of walking, running, crawling, step climbing, and shuffling. Overall, the distance traveled by the test subject during data collection was approximately 1,025 m. To reiterate, we labelled our dataset at every timestep in each motion trial with a binary midstance indicator that was generated from the optimal zero-velocity detector output.

\begin{table}[]
	\centering
	\caption{Zero-velocity labelling results. ``Num. Trials'' indicates the number of trials in which the detector achieved the lowest ARMSE compared to the others.}
	\label{tab:data-collect}
	\begin{tabular}{c  c  c  c c}
		\begin{tabular}[c]{@{}c@{}}Detector\end{tabular} & \begin{tabular}[c]{@{}c@{}}Avg.\\ Error (m)\end{tabular}  & \begin{tabular}[c]{@{}c@{}}Min.\\ Thresh.\end{tabular} & \begin{tabular}[c]{@{}c@{}}Max.\\ Thresh.\end{tabular} & \begin{tabular}[c]{@{}c@{}}Num.\\ Trials\end{tabular} \\ \midrule \T \T \B
		VICON & 0.074          & $2.25\times10^{-2}$ & $8.25\times10^{-1}$ & 15          \\
		SHOE  & \textbf{0.068} & $4.75\times10^5$    & $6.50\times10^8$    & \textbf{30} \\
		AMVD  & 0.336           & $1.00\times10^{-3}$ & $1.95\times10^{0}$  & 0           \\
		ARED  & 0.075          & $1.25\times10^{-2}$ & $2.70\times10^{0}$  & 13          \\
		MBGTD & 0.329          & $5.75\times10^{-3}$ & $9.75\times10^{-1}$ & 2           \\ \midrule \T \T \B
	\end{tabular}
	\vspace{-1.0cm}
\end{table}

\Cref{tab:data-collect} summarizes our midstance labelling results for all of the VICON training/testing trials. We see that the optimized SHOE detector was the most accurate overall, followed by VICON-based zero-velocity detection and then by the ARED. Note that our VICON detector was considered to be optimal for all of the non-walking motions, such as shuffling, crawling, and step climbing; the addition of this detector was critical to produce a training set that allowed the learned detector to generalize to non-walking motions. Columns 3--4 of the table indicate the smallest and largest optimized thresholds for each detector and illustrate how much the optimal threshold changes when a user's motion type or motion intensity changes. 

In practice, the accuracies achieved by these optimized detectors would never be reached if  a single threshold value was used during dynamic motions. However, our LSTM-based detector, having been trained with the midstance outputs from these optimized detectors, could in theory produce INS estimates with this level of accuracy. We show in \cref{sec:vicon-results} that, in fact, we come very close to achieving this accuracy.

\subsection{LSTM Network Architecture}
\label{sec:lstm-architecture}
We evaluated several network structures, including feedforward networks and recurrent neural network (RNN) variants (gated recurrent units and LSTMs). By treating an IMU measurement sequence as an ``IMU image,'' it would also be possible to apply CNNs for detection of midstance.  While we omit a full comparison from the current paper, in practice we found that an LSTM architecture achieved the highest accuracy for our task. An LSTM cell builds upon the baseline recurrent cell, which combines the current input $\mbf{x}^{(t)}$ with the network's hidden state $\mbf{h}^{(t-1)}$:
\begin{align}
\mbf{h}^{(t)} = \phi(\mbf{W}^{hx}\mbf{x}^{(t)} + \mbf{W}^{hh}\mbf{h}^{(t-1)}).
\end{align}
%
\noindent The output $\mbf{h}^{(t)}$ is a linear combination of the previous hidden state and the current input, passed through an activation function, $\phi(\cdot)$, such as $\tanh(\cdot)$. The matrices $\mbf{W}^{hx}$ and $\mbf{W}^{hh}$ contain the weights that are updated during training.

In addition to the hidden state $\mbf{h}^{(t)}$, LSTMs propagate an internal state $\mbf{s}^{(t)}$.  At each timestep, updates to $\mbf{s}^{(t)}$ are governed by two gate structures: the input gate $\mbf{i}^{(t)}$ and the forget gate $\mbf{f}^{(t)}$.  The input gate controls what elements of the input node $\mbf{g}^{(t)}$ are added to the state, while the forget gate removes (forgets) elements of the state that are no longer needed.  A third gate $\mbf{o}^{(t)}$ is used to choose the elements of $\mbf{s}^{(t)}$ that will be passed into $\mbf{h}^{(t)}$,
\vspace{-1.0mm}
\begin{align*}
\mbf{g}^{(t)} &= \phi(\mbf{W}^{\text{gx}}\mbf{x}^{(t)} + \mbf{W}^{\text{gh}}\mbf{h}^{(t-1)} + \mbf{b}_\text{g}),\\
\mbf{i}^{(t)} &= \sigma(\mbf{W}^{\text{ix}}\mbf{x}^{(t)} + \mbf{W}^{\text{ih}}\mbf{h}^{(t-1)} + \mbf{b}_\text{i}),\\
\mbf{f}^{(t)} &= \sigma(\mbf{W}^{\text{fx}}\mbf{x}^{(t)} + \mbf{W}^{\text{fh}}\mbf{h}^{(t-1)} + \mbf{b}_\text{f}),
\end{align*}

\begin{align*}
\mbf{o}^{(t)} &= \sigma(\mbf{W}^{\text{ox}}\mbf{x}^{(t)} + \mbf{W}^{\text{oh}}\mbf{h}^{(t-1)} + \mbf{b}_\text{o}),\\
\mbf{s}^{(t)} &= \mbf{g}^{(t)}\odot \mbf{i}^{(t)} + \mbf{s}^{(t-1)} \odot \mbf{f}^{(t)},\\
\mbf{h}^{(t)} &= \phi(\mbf{s}^{(t)}) \odot \mbf{o}^{(t)},
\end{align*}
where $\sigma(\cdot)$ is the sigmoid function and $\odot$ represents element-wise multiplication.  There are eight sets of weight parameters $\mbf{W}^{jk}$, where $j\in\{g,i,f,o\}$ and $k \in\{x,h\}$. The biases $\mbf{b}_g$, $\mbf{b}_i$, $\mbf{b}_f$, $\mbf{b}_o$ are also trainable network parameters.

LSTMs are a popular choice due to their resilience to the vanishing gradient problem, as $\mbf{s}^{(t)}$ provides a direct connection between timesteps that are temporally far apart.  Thus, gradients can be backpropagated through time over long sequences, enabling the network to learn long-range dependencies. For a more detailed explanation of LSTM networks, we refer the reader to \cite{Lipton:2015}.

Our zero-velocity detector consisted of a 6-layer LSTM, with 80 units per layer. We include a single fully-connected layer after the LSTM, which reduces the network's output to 2D. A softmax function is used to constrain the outputs to sum to 1, with the individual outputs corresponding to the detector's confidence that the IMU is in motion or stationary. We filter our output such that we only assume the IMU is stationary when the confidence is over 0.85 (empirically determined to minimize error caused by false-positive detections).

\begin{figure}[t]
	\centering
	\includegraphics[width=0.9\columnwidth]{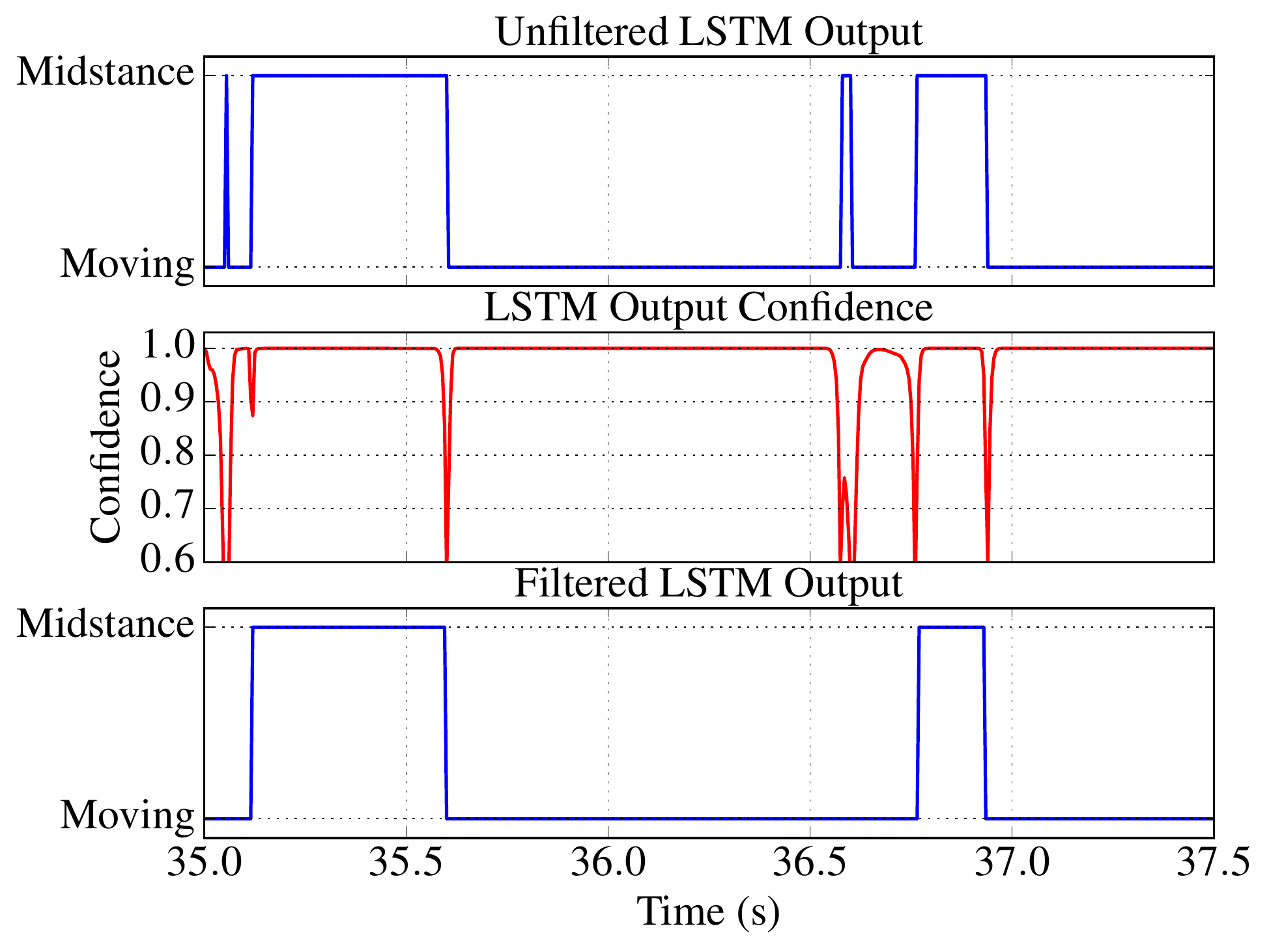}
	\caption{Unfiltered (top) and filtered (bottom) LSTM output. Filtering the LSTM output using its confidence (middle) removes the false-positive detections at 35.1 and 36.6 seconds.}
	\label{fig:zv-filter}
	\vspace{-0.5cm}
\end{figure}

\Cref{fig:zv-filter} illustrates the effects of removing low-probability zero-velocity predictions. False-positive detections are problematic as zero-velocity `events' occurring during movement will result in irreversible error accumulation.

\subsection{Training}
\label{sec:lstm-training}

The dataset described in \Cref{sec:datacollection2} was separated into a training set and a test set, consisting of 51 and 9 motion trials, respectively.  The total distance travelled in the training and test set were approximately 840 m and 182 m, respectively. Our LSTM input sequences consisted of raw, six-channel inertial data. From each motion trial, we extracted 7,000 clusters that each consisted of 100 consecutive IMU timesteps (or 0.5 seconds of inertial data at 200 Hz), resulting in a dataset size of $N=420,000$ individual samples overall. Each training sample $\mbf{x}_i \in\mathds{R}^{100\times6}$ had a single label, $y_i \in \{0,1\}$, that corresponded to the midstance label at the final timestep.

Similar to image augmentation methods used to increase the amount of data available for training CNNs \cite{Krizhevsky:2012}, Um et al.\ \cite{Kulic:2017} recently proposed several data augmentation methods for wearable sensors, and showed that augmentation significantly improved classification performance (for inertial data). We adopted three techniques from \cite{Kulic:2017} (rotation, scaling, and jittering) to improve our model's invariance to factors such as IMU placement/orientation, shoe type, and gait type. For each training sample, we applied a random \SO3 rotation $\mbf R$ (applied to $\mbf{a}$ and $\boldsymbol{\omega}$ for all data points in the sample), a random scaling factor, $s \in [0.92,1.02]$, to simulate faster or slower movement, and added zero-mean Gaussian noise $\mbf{n}$ ($\sigma=0.075$) to each channel. We trained our model for 300 epochs using the Adam optimizer \cite{Kingma:2014} with a standard cross-entropy loss function
\begin{equation}
\mathscr{L} = -\frac{1}{N}\sum_i^N y_i\log(p_i) + (1-y_i)\log(1-p_i).
\end{equation}
We used gradient clipping (limiting our gradient magnitude to be less than one) to avoid issues with exploding gradients. Our training parameters included a learning rate of $5\times10^{-3}$ (with its size reduced by half every 30 epochs), weight decay of $1\times10^{-5}$, and minibatch sizes of 800.  Our model was implemented in PyTorch \cite{paszke:2017} and training was carried out on an NVIDIA Titan X GPU.  We achieved classification accuracies of 97.2/97.0\% on the training and test sets, respectively.  We tuned our network hyperparameters by maximizing the test set accuracy, and present results on a validation set in \Cref{sec:results}.

At test time, we do not apply any transformations to the data. A single forward pass of a motion trial through the network efficiently outputs all midstance events for the whole trial, while propagating the hidden state throughout.

\section{System Overview}
\label{sec:overview}

We use an error-state (extended) Kalman filter (ESKF) for motion estimation, which separates our estimated state into a nominal state and a linearized error state. Our state consists of the IMU's position ($\mbf p_k$), velocity ($\mbf v_k$), and orientation in quaternion form ($\mbf q_k$),
\begin{align}
\label{eq:state}
\mathbf{x}_k &= \bbm \mbf p^{T}_k & \mbf v^{T}_k &  \mbf q^{T}_k \ebm^T.
\end{align}
A nonlinear motion model $\mbf{f}(\cdot)$ propagates the nominal state through time, by integrating the IMU's outputs $\{\mathbf{a}_k, \boldsymbol{\omega}_k\}$ (representing the three-axis linear accelerations and angular velocities expressed in the IMU frame) \cite{Nilsson:2014},
\begin{align}
	\label{eq:naveq}
	\mbf{x}_k = \bbm
	\mbf{p}_k\\
	\mbf{v}_k\\
	\mbf{q}_k\\ \ebm 
	= 
	\bbm
	\mbf{p}_{k-1} + \mbf{v}_{k-1}\Delta t \\ 
	\mbf{v}_{k-1} + \left( \mbf{R}(\mbf{q}_{k-1})\mbf{a}_k - \mbf{g} \right)\Delta t  \\
	\mbf{\Omega} (\boldsymbol{\omega}_k \Delta t)\mbf{q}_{k-1} \\
	\ebm,
\end{align}
\noindent where $\Delta t$ is the sampling period, $\mathbf{R}(\cdot) $ converts the orientation from a quaternion representation to a rotation matrix, $\mathbf{g}$ is the local gravity vector, and $\mbf{\Omega}(\cdot)\in\mathds{R}^{4\times4}$ is a linear update to the quaternion orientation that incrementally rotates $\mbf{q}_{k-1}$ by the integrated angular rate (computed through a backward, zeroth-order integration).  The nominal state accumulates error over time, which we estimate separately with our linearized error state that incorporates noise and perturbations to the system.

Upon detection of a zero-velocity event, the velocity error of the system is rendered observable, as the known velocity is (approximately) zero. This linear measurement is fused with the error state via the standard EKF update step. After adjusting the nominal state based on the error, the error state is reset to zero. We note that zero-velocity measurements are able to correct for errors in all filter states except for the yaw angle \cite{Foxlin:2005}. We refer the reader to \cite{Foxlin:2005,Sola:2016} for a more detailed explanation of the ESKF.

We note that our system is implemented in a minimalistic manner: the literature is rich with various improvements to the baseline zero-velocity-aided INS. Barometer and magnetometer measurements \cite{Bird:2011, Yang:2017} are often incorporated as well, since these sensors are built into many IMUs. Several map-aided implementations are also available, which globally bound error accumulation by probabilistically sampling trajectories and removing those that pass through walls \cite{Widyawan:2008}.  Multi-IMU implementations improve localization accuracy by imposing additional constraints that the distances between multiple body-mounted IMUs cannot exceed a maximum distance \cite{Prateek:2013, Tjhai:2017}. While these methods further improve localization accuracy, the emphasis of our paper is on zero-velocity detection, and comparison of our learned detector with existing detectors is most clear when using a baseline system.

\section{Inertial Localization Experiments}
\label{sec:results}

We describe two quantitative evaluations of our system: \Cref{sec:vicon-results} presents the results using the dataset collected within our VICON tracking area, while \Cref{sec:hallway-results} presents results using data collected throughout the University of Toronto Institute for Aerospace Studies (UTIAS). Finally, in \Cref{sec:motions-results}, we qualitatively demonstrate how our learned zero-velocity detector is highly robust to varying motion types. We note that, for all of the results discussed, we used a single learned model (with fixed parameters) that produced the highest zero-velocity detection accuracy on our VICON test set (i.e., the model with the best capability of generalizing to new data outside of the training set).

\subsection{VICON Laboratory Trials}
\label{sec:vicon-results}

\begin{figure}[t]
	\small
	\centering
	\includegraphics[width=0.48\textwidth]{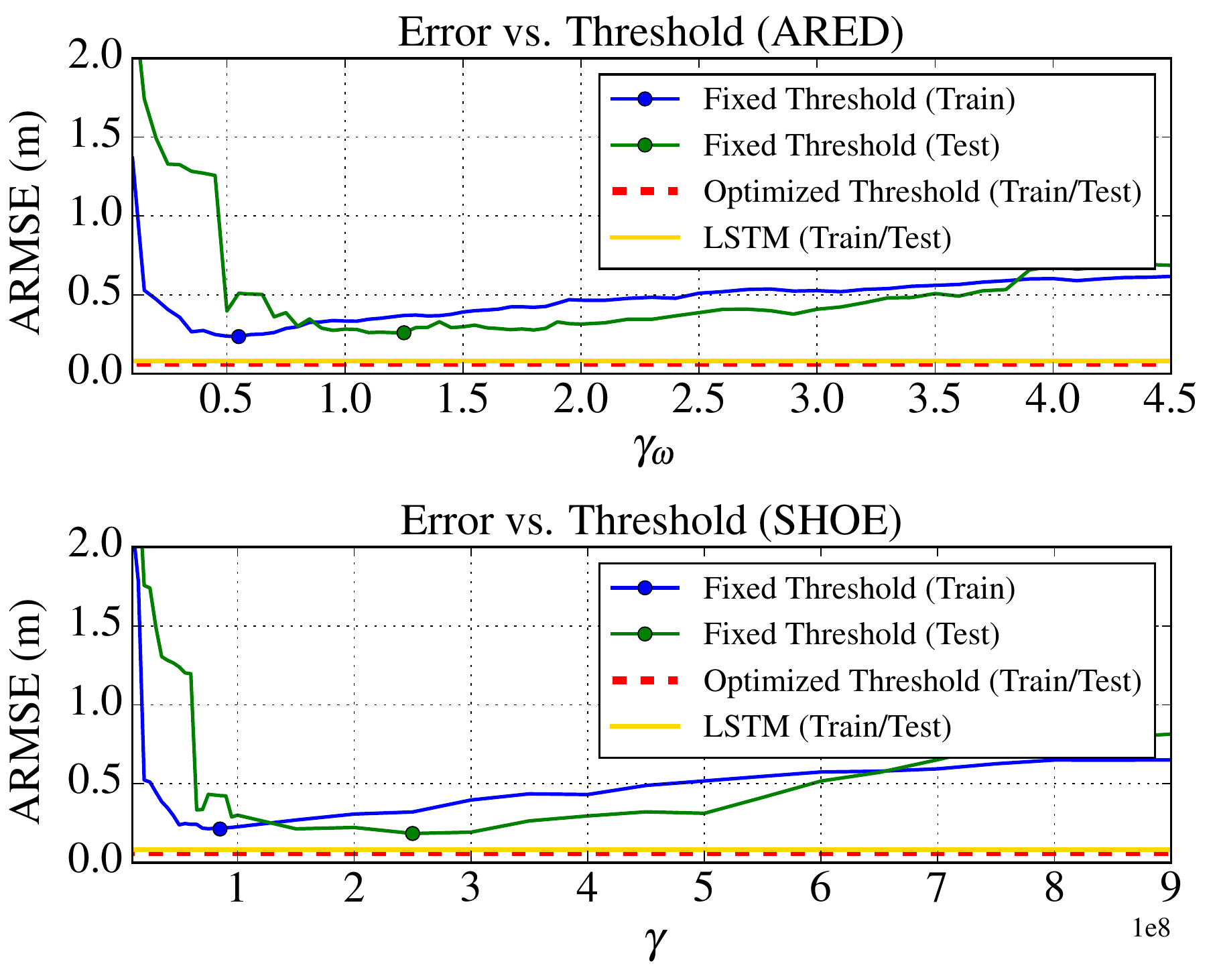}
	\caption{Error vs. threshold for all of the VICON motion trials. Our LSTM-based detector closely approximates the optimized-threshold ``ground truth,'' and outperforms the fixed-threshold SHOE or ARED.}
	\label{fig:error-vs-thresh}
	\vspace{-0.1cm}
\end{figure}

For our motion trial dataset consisting of the training and test sets described in \Cref{sec:datacollection1} and \Cref{sec:datacollection2}, we evaluated the INS position error (ARMSE) while using SHOE, ARED, and our LSTM-based model for zero-velocity detection. \Cref{fig:error-vs-thresh} depicts how the choice of threshold affects the ARMSE. Reinforcing our criticism of fixed-threshold detection, we see that using fixed thresholds for a set of motion trials produces errors that are significantly larger than the error produced while using the optimized per-trial threshold. Furthermore, we see that our LSTM-based zero-velocity detector yields position errors that are close to the error produced from the optimized per-trial threshold detectors. We emphasize that this significant reduction in error is achieved \emph{without} the need to tune any parameters during operation.

\begin{figure}[t]
	\vspace{0.1cm}
	\small
	\centering
	\includegraphics[width=0.75\columnwidth]{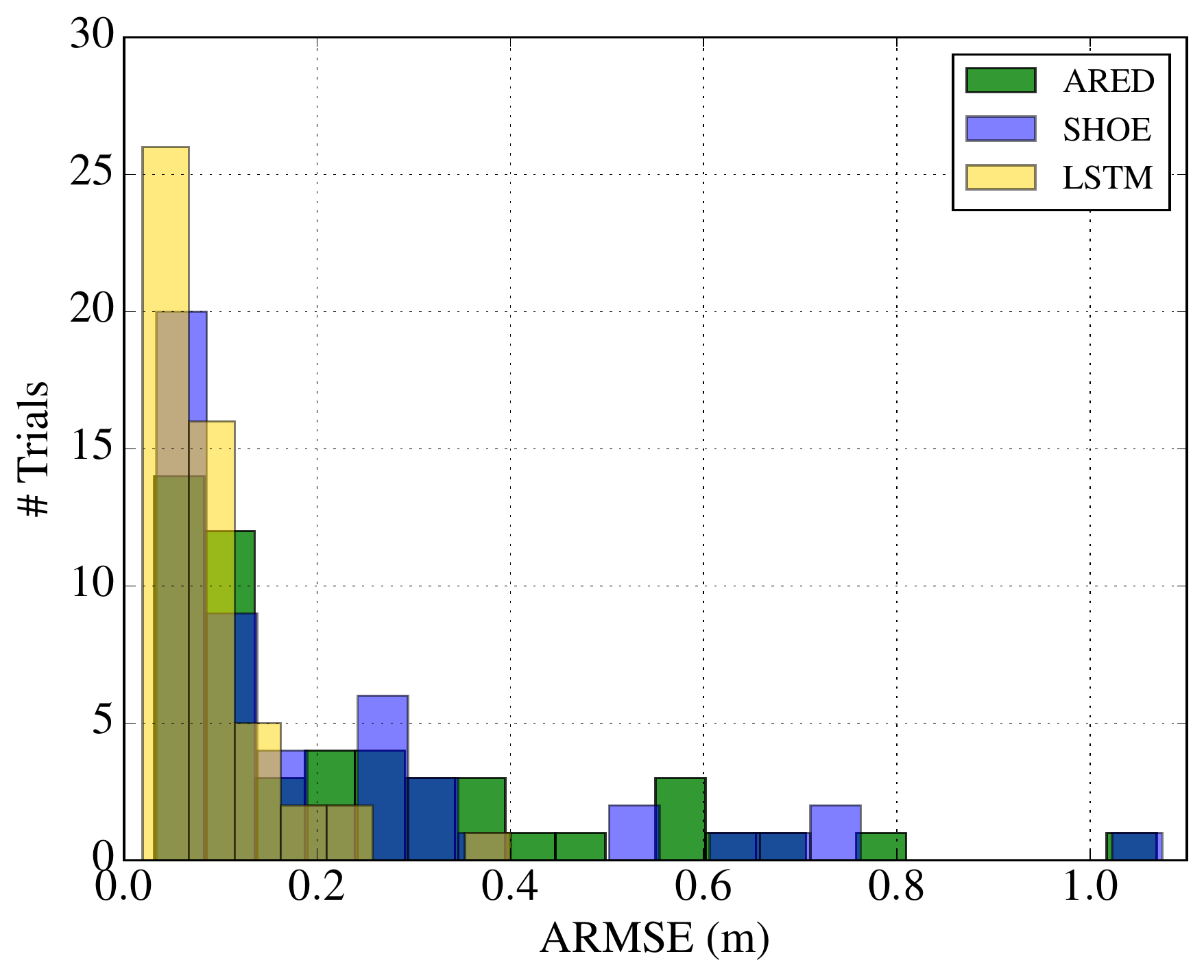}
	\caption{Error distribution for the VICON training set trials, when using the fixed-threshold ARED or SHOE detector, compared with our LSTM-based zero-velocity detector.}
	\label{fig:vic-hist}
	\vspace{-0.1cm}
\end{figure}

\Cref{tab:vicon-results} presents the same results in a tabular format. We highlight the optimal training and test set errors using fixed thresholds with SHOE and ARED, and compare these with the LSTM-based detection, noting that our detector produced a $60.1\%$ and $51.6\%$ reduction in error on the training and test sets, respectively. The error reduction confirms that our learned model has generalized beyond the training data.  Additionally, we demonstrate that the LSTM accuracy approaches that of the optimized detectors (which is the upper limit that would be achieved if the LSTM was trained to 100\% accuracy). \Cref{fig:vic-hist} depicts the error distribution for our VICON training dataset. We see that there are a significant number of trials where fixed-threshold detection results in large error, due to the existence of a subset of trials where the current threshold greatly differs from that trial's optimal threshold. Contrary to this, our LSTM-based detector error is distributed around a lower mean, with fewer outliers.

\begin{table}[]
	\begin{threeparttable}
	\centering
	\caption{2D Localization error (ARMSE) for all of the motion trials within the VICON motion capture area. Errors are computed by comparing the INS estimate with the VICON ground truth position data.}
	\label{tab:vicon-results}
	\begin{tabular}{cccccc}
		\textbf{Detector}                                             & \textbf{}         & \multicolumn{2}{c}{\textbf{Training Set Error (m)}} & \multicolumn{2}{c}{\textbf{Test Error (m)}} \\ \midrule \T \T \B
		\textbf{}                                                     & \textbf{$\gamma$} & Avg.                     & End                      & Avg.                 & End                  \\ \midrule \T \T \B
		\multirow{7}{*}{SHOE}                                         & 6e7               & 0.242                    & 0.296                    & 1.197                & 1.907                \\
		& 8e7               & 0.217                    & 0.302                    & 0.428                & 1.100                \\
		& 8.5e7$^1$             & \textbf{0.213}           & \textbf{0.281}           & 0.424                & 1.086                \\
		& 1e8               & 0.227                    & 0.297                    & 0.300                & 0.815                \\
		& 2.5e8$^2$              & 0.321                    & 0.345                    & \textbf{0.184}       & \textbf{0.244}       \\
		& 3e8               & 0.397                    & 0.429                    & 0.193                & 0.293                \\
		& 5e8               & 0.517                    & 0.567                    & 0.312                & 0.458                \\ \midrule \T \T \B
		\multirow{7}{*}{ARED}                                         & 0.1               & 1.367                    & 1.41                     & 2.454                & 2.744                \\
		& 0.5               & \textbf{0.237}           & \textbf{0.359}           & 0.397                & 0.724                \\
		& 0.55$^1$               & 0.235                    & 0.344                    & 0.509                & 1.218                \\
		& 1                 & 0.333                    & 0.428                    & \textbf{0.282}       & \textbf{0.413}       \\
		& 1.5               & 0.391                    & 0.468                    & 0.297                & 0.497                \\
		& 2                 & 0.465                    & 0.547                    & 0.314                & 0.454                \\
		& 2.5$^2$                & 0.510                    & 0.567                    & 0.387                & 0.496                \\ \midrule \T \T \B
		\begin{tabular}[c]{@{}c@{}}Optimal\\ Detection\end{tabular} & -                 & \textbf{0.068}           & \textbf{0.079}           & \textbf{0.075}       & \textbf{0.057}       \\ \midrule \T \T \B
		LSTM (ours)                                                   & -                 & \textbf{0.085}           & \textbf{0.108}           & \textbf{0.089}       & \textbf{0.077}       \\ \bottomrule
	\end{tabular}
	\begin{tablenotes}
		\small
		\item 1.  Threshold resulted in the lowest ARMSE over the training set.
		\item 2.  Threshold resulted in the lowest ARMSE over the test set.
	\end{tablenotes}
	\end{threeparttable}
	\vspace{-0.5em}
\end{table}

\subsection{Building Trials}
\label{sec:hallway-results}

\begin{figure*}[t]
	\centering
	\begin{subfigure}[]{0.32\textwidth}
		\includegraphics[width=\textwidth]{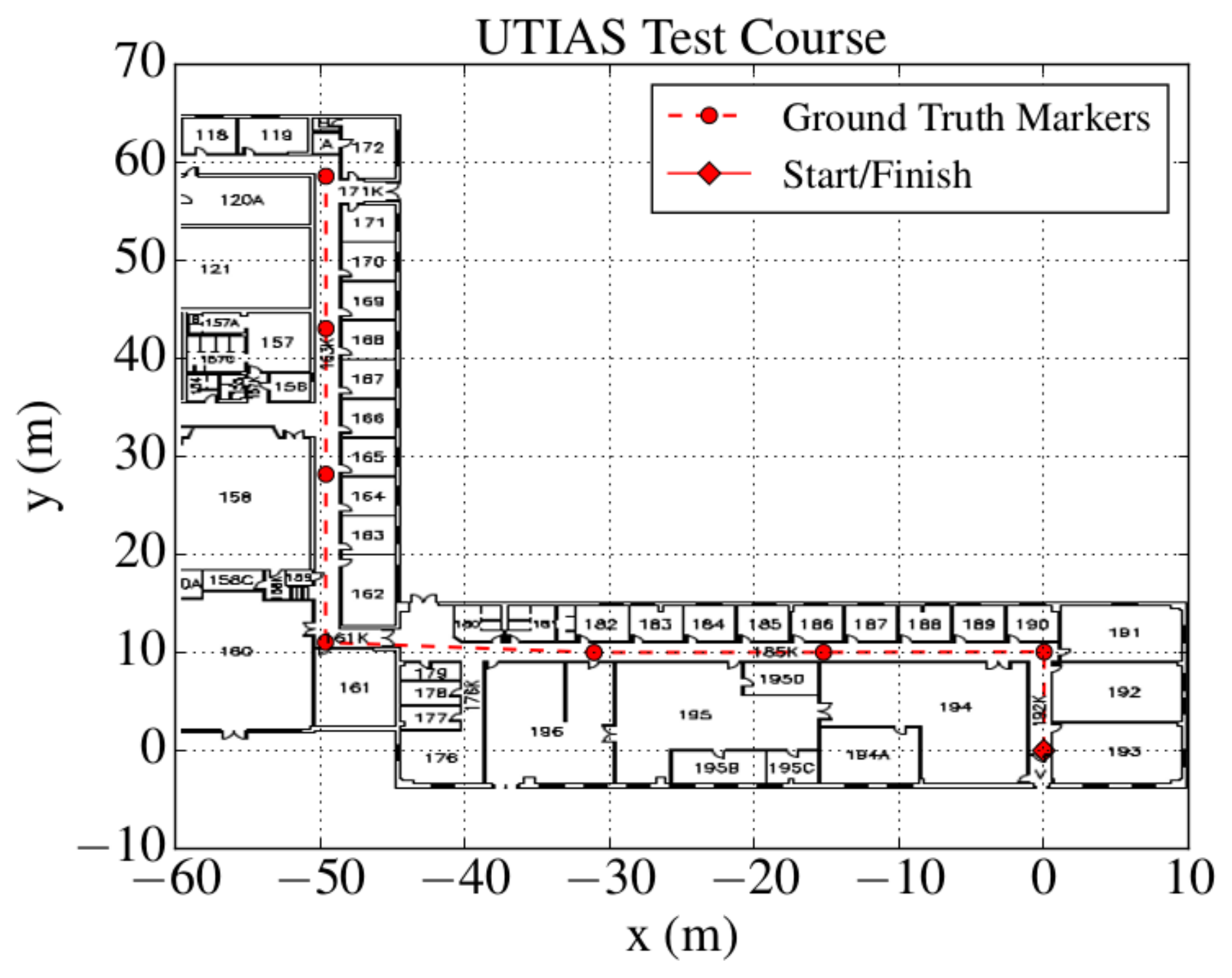}
		\caption{UTIAS test course.}
		\label{fig:hallway}
	\end{subfigure}
	~
	\begin{subfigure}[]{0.32\textwidth}
		\includegraphics[width=\textwidth]{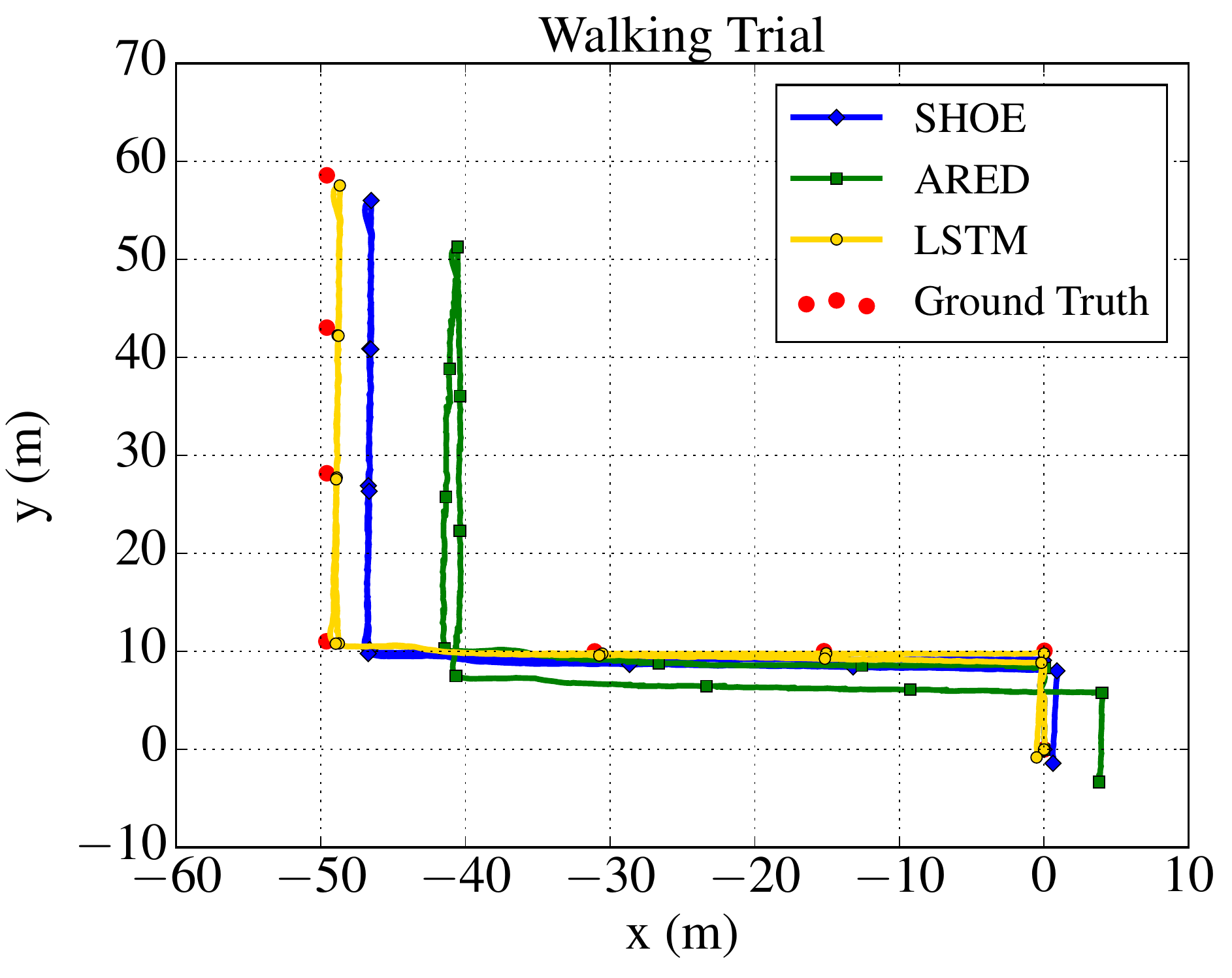}
		\caption{Walking motion trial.}
		\label{fig:hallwalk}
	\end{subfigure} 
	\begin{subfigure}[]{0.32\textwidth}
		\includegraphics[width=\textwidth]{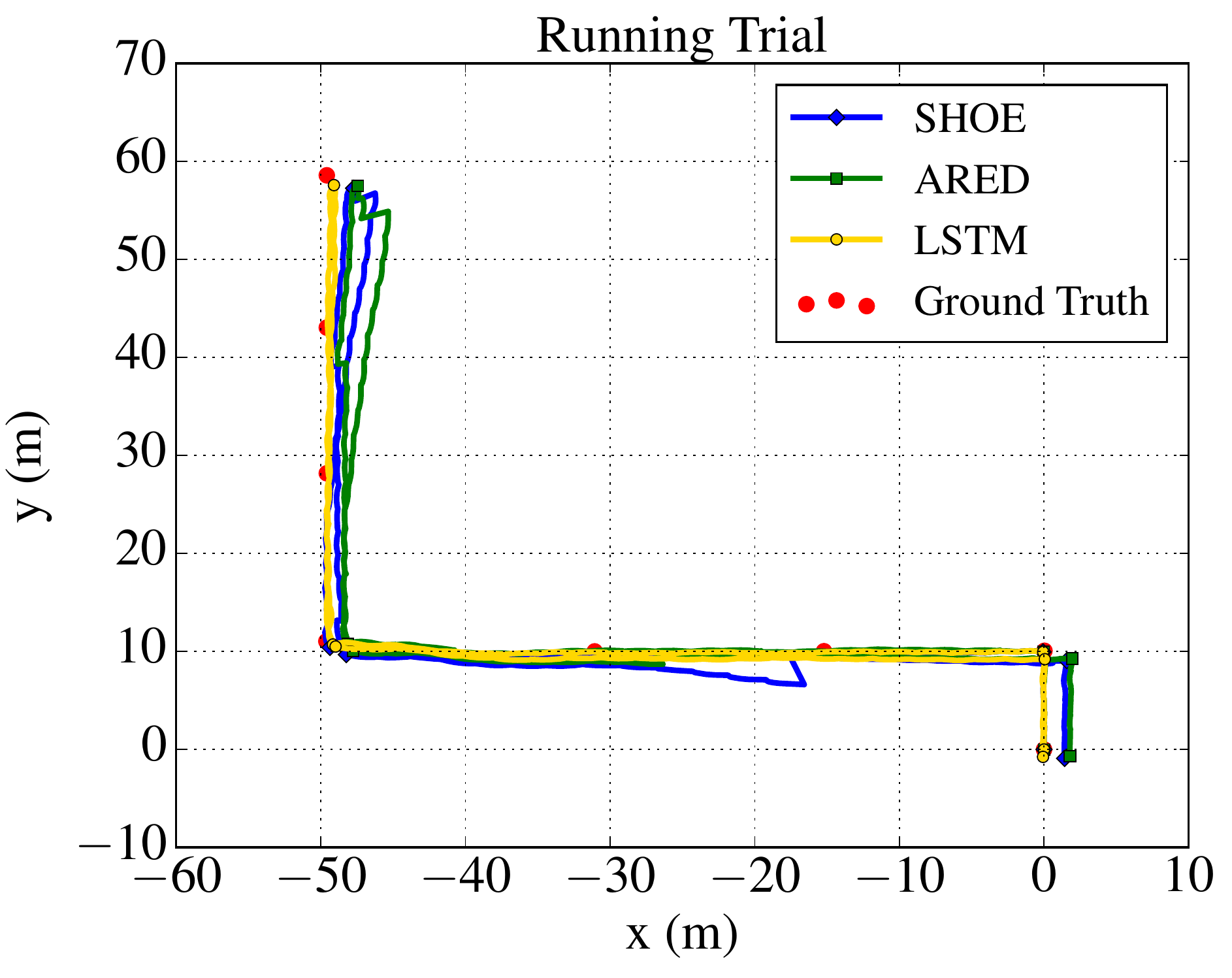}
		\caption{Running motion trial.}
		\label{fig:hallrun}
	\end{subfigure} \\
	~
	\caption{Comparing INS trajectories for walking and running motions.  For both motions, the ARED and SHOE detector used the optimal thresholds ($\gamma_{shoe}=8.5\text{e}7$, $\gamma_{ared}=0.55$) that resulted in minimal errors in the VICON training set.}
	\label{fig:halltraj}
	\vspace*{-2mm}
\end{figure*}

Rather than limiting ourselves to the small area where VICON ground truth is available, we collected data from users moving throughout the halls and rooms of a campus building. Similar to our prior work \cite{Wagstaff:2017}, we established ground truth along our test trajectory at a series of discrete points (floor markers) spaced throughout the building, as seen in \Cref{fig:hallway}. The ground truth markers were surveyed in using a Leica Nova MS50 MultiStation: six points on each marker were measured to determine the marker's pose in a local coordinate frame, and an analytic point cloud alignment algorithm \cite{Umeyama1991-ws} was used to transform the marker coordinates into a global reference (navigation) frame. When approaching each marker, test subjects attempted to step on the marker's origin, while pressing a handheld trigger that recorded the current timestamp. Consequently, the IMU pose at these timestamps could be compared with the known ground truth marker positions to evaluate the position error. The hallway trajectory length was approximately 220 m (round-trip). We used a VectorNav VN-100 IMU operating at 200 Hz for the building trials, illustrating how our learned model is able to generalize to new IMUs at test time.

Each test subject performed three types of motion trials that consisted of repetitions of walking, running, and combinations of these dynamic motions. The walking and running trials were performed exclusively in the hallways, while users additionally entered two supplementary rooms during the walk/run trials.  In the first room, subjects were instructed to alternate between walking and running. In the second room (a lecture hall) subjects were required to walk up a set of stairs (six steps that rose to a height of one meter), and down a long ramp which brought the test subject back to floor level. The RMSE was evaluated at all eight markers for the walking and combined walk/run trials, while the running trials were only evaluated at four markers (located at the corners of the trajectory) to prevent the test subjects from slowing down more frequently. In total, we collected over 6.6 km of motion data during the building trials.

\Cref{tab:hall-results} presents the ARMSE (over all marker locations) from all trials that were collected. We compare the results from baseline methods (SHOE or ARED using fixed thresholds) with our proposed LSTM-based zero-velocity detector. We see that the LSTM detector consistently outperforms the ARED and SHOE detector, producing an average error reduction of 34.8\% compared to the best-performing threshold from a baseline detector. \Cref{fig:hallwalk,,fig:hallrun} illustrates these results for a single walking and running trial.  Our detector is able to avoid the catastrophic failures that are evident when improper thresholds are used. 
\begin{figure*}[t]
	\centering
	\begin{subfigure}[]{0.32\textwidth}
		\includegraphics[width=\textwidth]{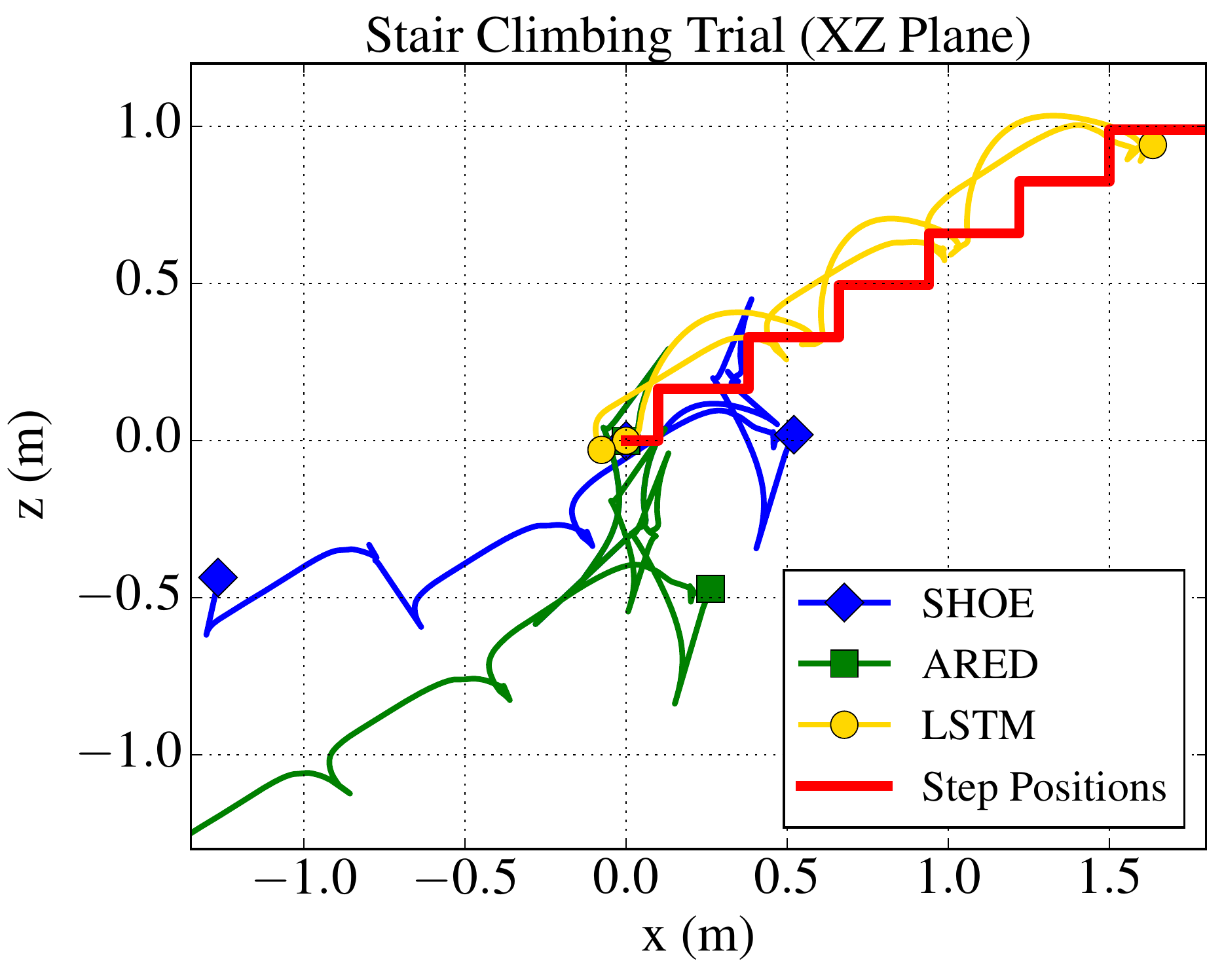}
		\caption{Stair Climbing.}
		\label{fig:stairs}
	\end{subfigure}
	~
	\begin{subfigure}[]{0.32\textwidth}
		\includegraphics[width=\textwidth]{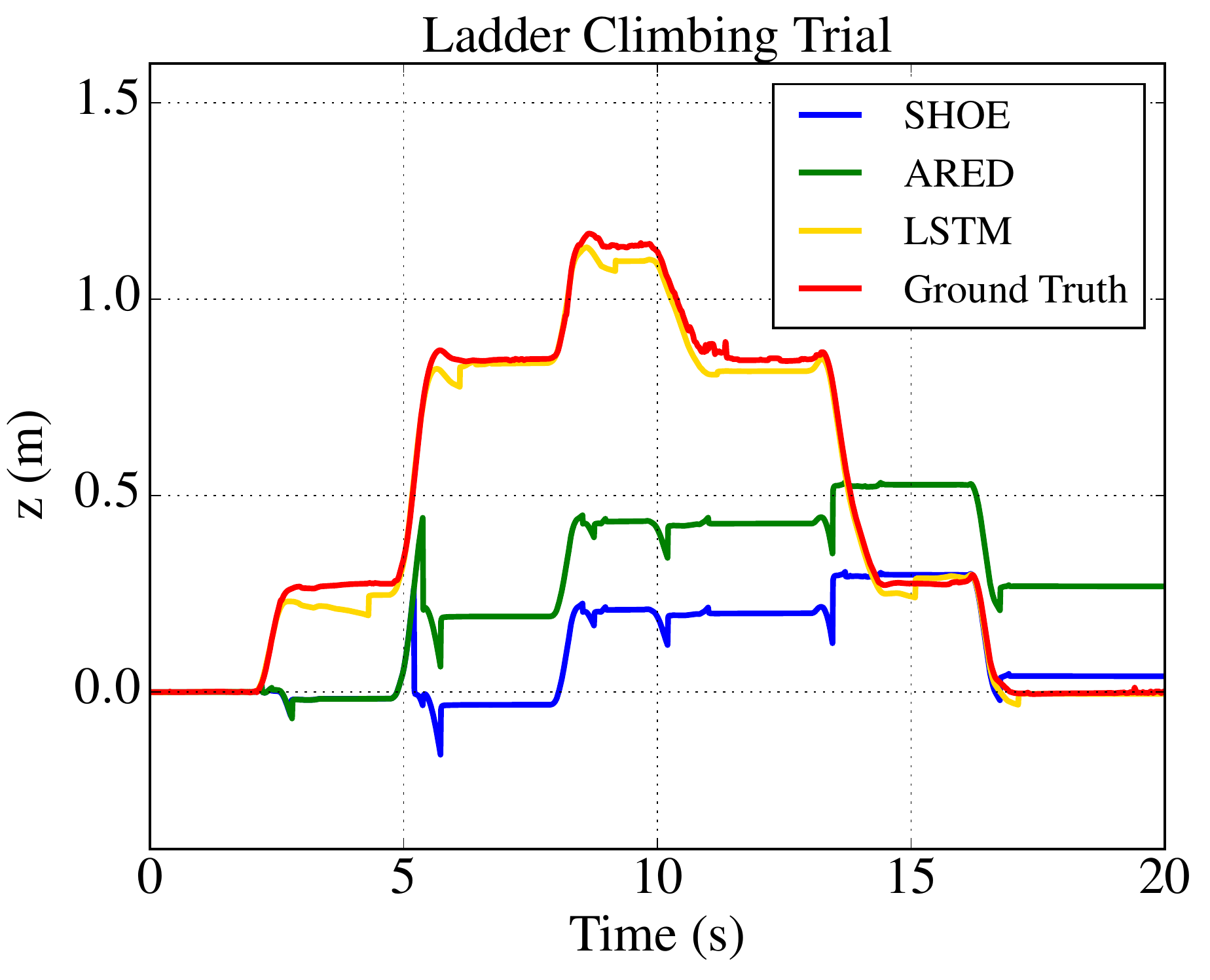}
		\caption{Ladder Climbing.}
		\label{fig:ladder}
	\end{subfigure} 
	\begin{subfigure}[]{0.32\textwidth}
		\includegraphics[width=\textwidth]{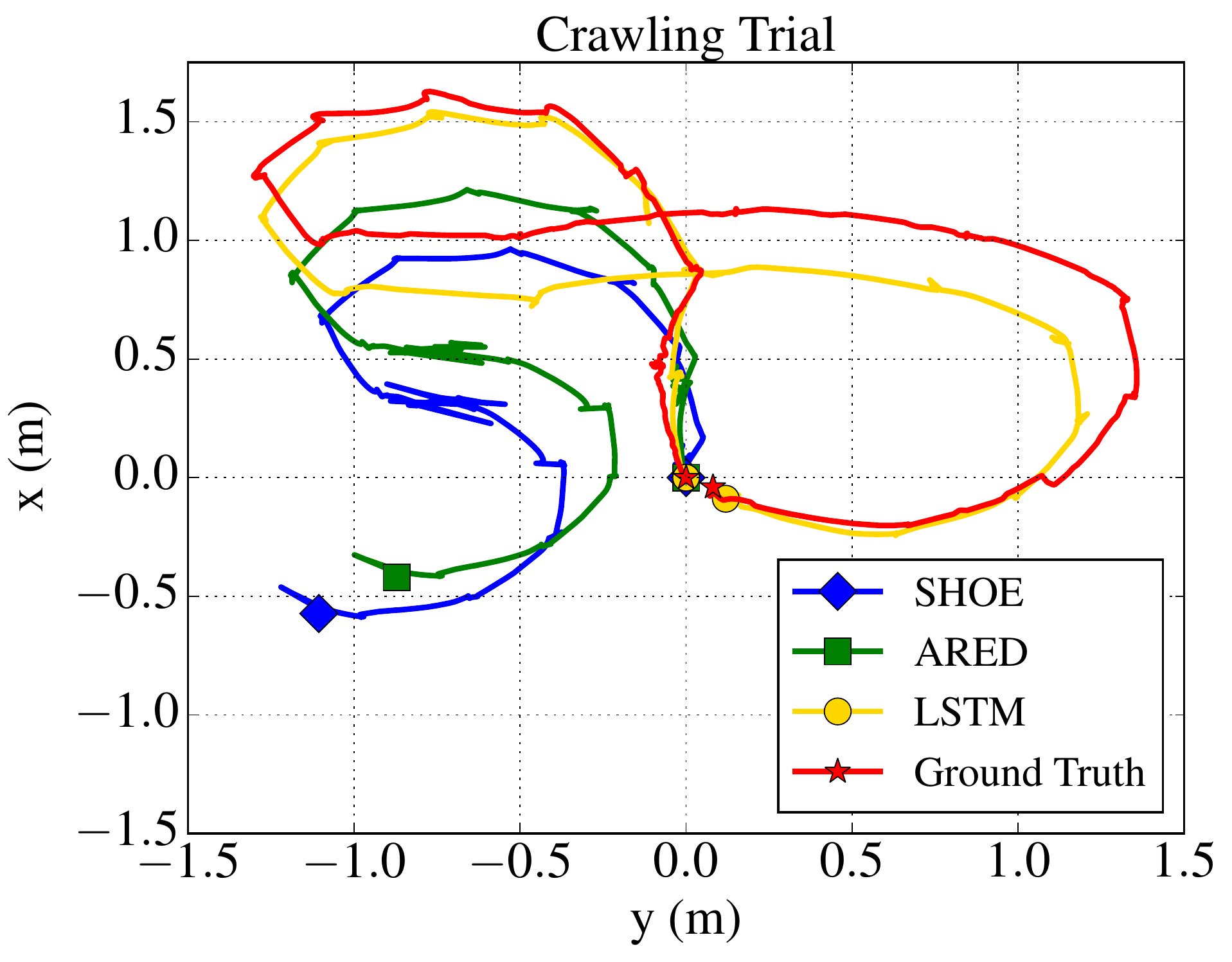}
		\caption{Crawling.}
		\label{fig:crawl}
	\end{subfigure}
	~
	\caption{Comparison of INS trajectories generated while using different zero-velocity detectors. We note that our LSTM zero-velocity detector is able to function effectively  during stair climbing, ladder climbing, and crawling motions, while two popular hand-crafted detectors fail.}
	\label{fig:motions}
	\vspace*{-2mm}
\end{figure*}
\begin{table}[]
	\centering
	\caption{Evaluation of the accuracy of various detectors on our hallway dataset. For the SHOE detector and the ARED, we used the optimal thresholds that were calculated for the VICON dataset.}
	\label{tab:hall-results}
	\begin{tabular}{c @{\hspace{0.5\tabcolsep}} c @{\hspace{0.5\tabcolsep}} c @{\hspace{0.9\tabcolsep}} c@{\hspace{0.9\tabcolsep}} c @{\hspace{0.9\tabcolsep}}c @{\hspace{0.9\tabcolsep}}c @{\hspace{0.9\tabcolsep}}c @{\hspace{0.9\tabcolsep}}c}
		\multicolumn{2}{c}{\textbf{Motion}} & \multicolumn{7}{c}{\textbf{ARMS Error (m)}} \\ \midrule \T \T \B
		Type & Subject & \multicolumn{3}{c}{ARED} & \multicolumn{3}{c}{SHOE} & LSTM \\ \midrule \T \T \B
		&  & \begin{tabular}[c]{@{}c@{}}$\gamma_{\text{walk}}$ \\ (0.3)\end{tabular} & \begin{tabular}[c]{@{}c@{}}$\gamma_{\text{mid}}$ \\ (0.55)\end{tabular} & \begin{tabular}[c]{@{}c@{}}$\gamma_{\text{run}}$ \\ (0.8)\end{tabular} & \begin{tabular}[c]{@{}c@{}}$\gamma_{\text{walk}}$\\ (1e7)\end{tabular} & \begin{tabular}[c]{@{}c@{}}$\gamma_{\text{mid}}$\\ (8.5e7)\end{tabular} & \begin{tabular}[c]{@{}c@{}}$\gamma_{\text{run}}$\\ (35e7)\end{tabular} &  \\ \midrule \T \T \B
		\multirow{5}{*}{Walking} & 1 & 1.525 & 3.152 & 4.542 & 0.575 & 1.354 & 5.914 & \textbf{0.550} \\
		& 2 & 1.216 & 1.487 & 1.862 & \textbf{0.921} & 1.250 & 2.168 & 0.931 \\
		& 3 & 2.754 & 2.935 & 3.165 & 2.462 & 2.873 & 3.818 & \textbf{2.181} \\
		& 4 & 1.447 & 1.549 & 1.874 & 1.197 & 1.235 & 3.155 & \textbf{1.158} \\
		& 5 & 0.674 & 0.778 & 0.881 & \textbf{0.568} & 0.792 & 1.293 & 0.669 \\ \midrule \T \T \B 
		\multirow{5}{*}{Running} & 1 & 0.937 & 1.121 & 1.913 & 1.037 & 0.760 & 2.452 & \textbf{0.666} \\
		& 2 & 1.875 & 2.164 & 2.269 & 19.153 & 1.886 & 0.997 & \textbf{0.785} \\
		& 3 & 1.666 & 1.536 & 1.672 & 1.902 & 1.569 & 2.642 & \textbf{1.045} \\
		& 4 & 0.902 & 1.619 & 1.953 & 1.255 & 1.550 & 1.627 & \textbf{0.764} \\
		& 5 & 1.327 & 17.461 & 18.544 & 1.503 & 1.394 & 1.541 & \textbf{1.097} \\ \midrule \T \T \B 
		\multirow{5}{*}{Combined} & 1 & 1.255 & 1.801 & 2.574 & 1.070 & 1.325 & 3.546 & \textbf{0.893} \\
		& 2 & 2.737 & 2.353 & 2.555 & 2.315 & 2.084 & 2.475 & \textbf{1.456} \\
		& 3 & 2.662 & 3.165 & 3.572 & 2.250 & 2.930 & 4.054 & \textbf{1.740} \\
		& 4 & 1.171 & 1.378 & 1.469 & 1.158 & 1.304 & 2.229 & \textbf{1.121} \\
		& 5 & 2.756 & 2.838 & 2.811 & 2.648 & 2.805 & 1.600 & \textbf{1.183} \\ \midrule \T \T \B 
		Mean &  & 1.660 & 3.022 & 3.444 & 2.668 & 1.674 & 2.634 & \textbf{1.083} \\ \bottomrule
	\end{tabular}
\end{table}

\subsection{Evaluation for Varying Motion Type}
\label{sec:motions-results}

Our final results show how our LSTM-based detector is able to accurately detect stationary periods of the IMU during motion types that substantially vary from walking or running. In \Cref{fig:motions}, we see that accurate trajectories are estimated for stair climbing, ladder climbing, and crawling motions while using our LSTM-based zero-velocity detector, whereas both the SHOE and ARED detectors fail. We attribute the breakdown of SHOE and ARED to their dependence on angular velocity (see \Cref{eq:shoe,,eq:ARED}). For motions such as stair climbing, ladder climbing, or crawling, the foot rotation rate is small, which results in false-positive zero-velocity measurements when using angular-velocity-dependent detectors. In contrast to this, our LSTM-based zero-velocity detector had been trained with labels from our linear-velocity-based VICON detector, which operates independently of angular velocity. We are therefore able to achieve accurate zero-velocity detection for these motion types.

\section{Conclusion \& Future Work}

We have implemented a learned zero-velocity detector that uses an LSTM network to identify stationary periods of an IMU from raw inertial measurements.  Our LSTM-based detector provides accurate zero-velocity measurements to a zero-velocity-aided INS, limiting error accumulation that results from the integration of noisy IMU readings. We demonstrated that our detector is able to generalize to varying gait types, users, shoe types, IMU types, and IMU placements on the foot. We validated the effectiveness of our learned model by showing that INS position estimates are more accurate with our detector compared to two other popular threshold-based zero-velocity detectors. Finally, we demonstrated that our LSTM-based zero-velocity detector is able to accurately detect stationary periods during a wide range of motions, such as walking, running, crawling, and stair and ladder climbing. In future work, we plan to investigate the use of IMU arrays to further improve upon our learned, single-IMU detector.

\begin{center} \subsection*{Acknowledgements} \end{center}
\small

This work was supported in part by the Natural Sciences and Engineering Research Council (NSERC) of Canada. We gratefully acknowledge the contribution from NVIDIA Corporation (through their Hardware Grant Program), who provided the Titan X GPU used for this research.

\balance
\bibliographystyle{IEEEtran}
\bibliography{2018-wagstaff-lstmzv-ipin.bib}
\end{document}